\documentclass[letterpaper, 9pt, conference]{ieeeconf}  

\IEEEoverridecommandlockouts                              

\usepackage{graphicx}
\usepackage{booktabs}  
\usepackage{multirow}
\usepackage{amssymb}
\usepackage{amsmath}
\newcommand{\eg}{\textit{e}.\textit{g}.}
\newcommand{\etal}{\textit{et al}.}
\newcommand{\ie}{\textit{i}.\textit{e}.}

\title{\LARGE \bf
Transformer-CNN Cohort: Semi-supervised Semantic Segmentation by the Best of Both Students
}

\author{Xu Zheng$^{1}$, \textit{IEEE Student Member}, Yunhao Luo$^{4}$, Chong Fu$^{3}$, Kangcheng Liu$^{1}$, \textit{IEEE Member}, \\ Lin Wang$^{1,2}$$^\dagger$, \textit{IEEE Member} 
\thanks{$^\dagger$Corresponding author}
\thanks{$^{1}$Xu zheng is with AI Thrust,
        HKUST(GZ), Guangzhou, China,
        Email: {\tt\small xzheng287@connect.hkust-gz.edu.cn}}%
\thanks{$^{4}$Y. Luo is with the Department of Computer Science, Brown University, USA,
        Email: {\tt\small yluo73@cs.brown.edu}}%
\thanks{$^{3}$C. Fu is with the Department of Computer Science and Engineering, Northeastern University, Shenyang, China,
        Email: {\tt\small fuchong@mail.neu.edu.cn}}%
\thanks{$^{1}$K. Liu is with SMMG/ROAS Thrust, HKUST(GZ),
        Email: {\tt\small kangchengliu@hkust-gz.edu.cn }} 
\thanks{$^{1,2}$L. Wang is with AI/CMA Thrust, HKUST(GZ) and Dept. of CSE, HKUST, China,
        Email: {\tt\small linwang@ust.hk}}%
}

\begin{document}

\maketitle
\thispagestyle{empty}
\pagestyle{empty}

\begin{abstract}
The popular methods for semi-supervised semantic segmentation mostly adopt a unitary network model using convolutional neural networks (CNNs) and enforce consistency of the model’s predictions over perturbations applied to the inputs or model. However, such a learning paradigm suffers from two critical limitations: 
a) learning the discriminative features for the unlabeled data; b) learning both global and local information from the whole image. 
In this paper, we propose a novel Semi-supervised Learning (SSL) approach, called Transformer-CNN Cohort (\textbf{TCC}), that consists of two students with one based on the vision transformer (ViT) and the other based on the CNN. Our method subtly incorporates the multi-level consistency regularization on the predictions and the heterogeneous feature spaces via pseudo-labeling for the unlabeled data. First, as the inputs of the ViT student are image patches, the feature maps extracted encode crucial class-wise statistics. 
To this end, we propose class-aware feature consistency distillation (\textbf{CFCD}) that first leverages the outputs of each student as the pseudo labels and generates class-aware feature (CF) maps for knowledge transfer between the two students. 
Second, as the ViT student has more uniform representations for all layers, we propose consistency-aware cross distillation (\textbf{CCD}) to transfer knowledge between the pixel-wise predictions from the cohort. We validate the TCC framework on Cityscapes and Pascal VOC 2012 datasets, which outperforms existing SSL methods by a large margin. 

\end{abstract}

\section{INTRODUCTION}
Semantic segmentation~\cite{long2015fully,chen4617170frozen} is a crucial scene understanding task in computer and robotic vision, aiming to generate pixel-wise category prediction of an image. Most of the state-of-the-art (SoTA) methods focus on exploring the potential of convolutional neural networks (CNNs) and learning strategies~\cite{ronneberger2015u,wang2020deep}. However, a hurdle of training these models is the lack of large-scale and high-quality annotated datasets, imposing much burden for real applications,~\eg, autonomous driving~\cite{yang2018denseaspp}. 
Consequently, growing attention has been paid to deep semi-supervised learning (SSL) for semantic segmentation \cite{chen2021semi} using the labeled data and additional unlabeled data.

\begin{figure}[h!]
\centering
\includegraphics[width=0.36\textwidth]{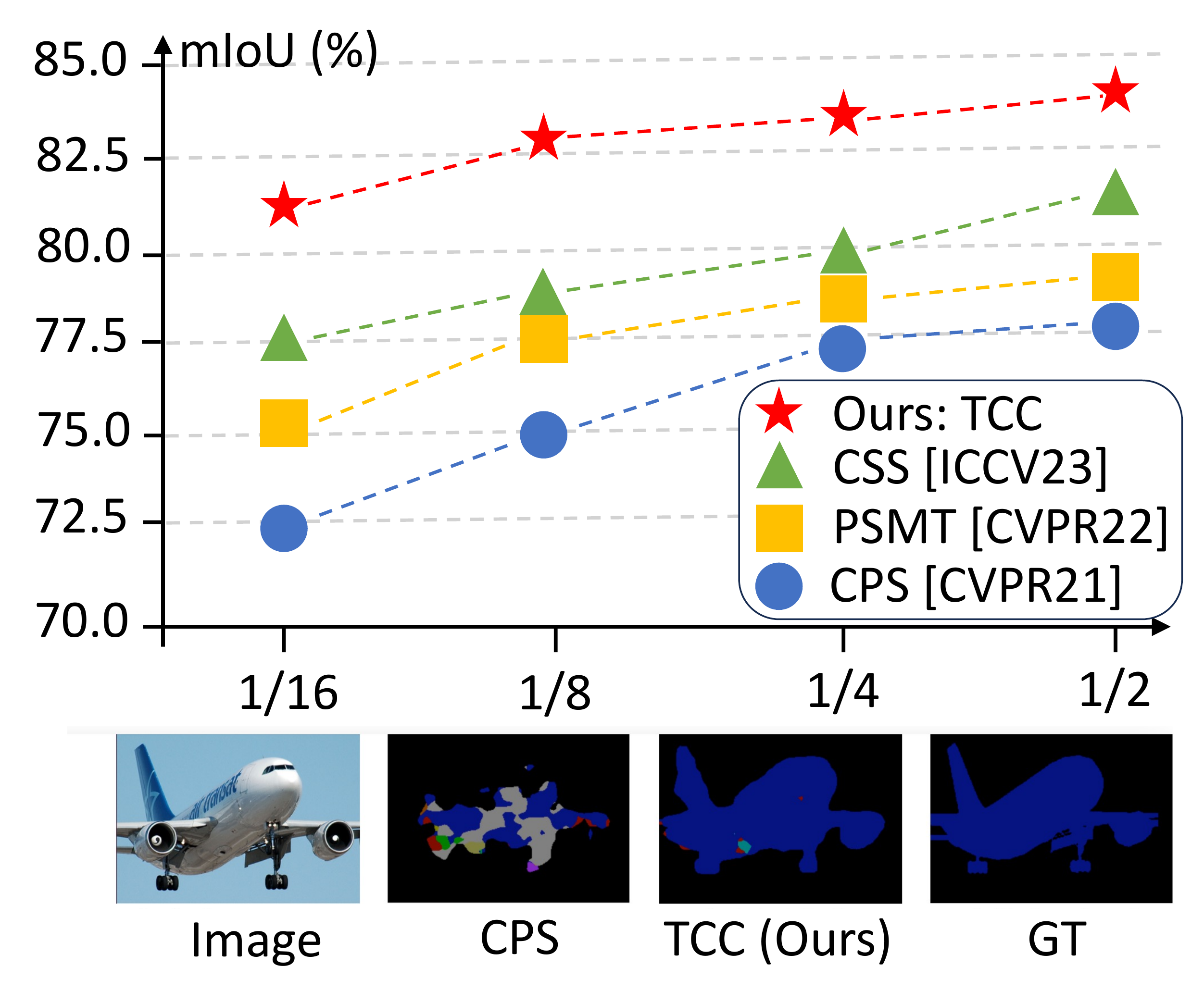}
\vspace{-10pt}
\caption{\textbf{Increment over SoTA baseline} on the \textit{PASCAL VOC} val set. Our TCC yields a noticeable improvement over previous methods, \eg, CPS~\cite{chen2021semi}. 
}
\label{Figure.1}
\vspace{-10pt}
\end{figure}

The dominant deep SSL methods rely on consistency regularization~\cite{tarvainen2017mean,ke2019dual}, pseudo labeling~\cite{ncps}, entropy minimization~\cite{grandvalet2005semi} and bootstrapping~\cite{grezl2013boostrapping}, etc. However, these methods are only limited to classification, and their applications to semantic segmentation are still restricted~\cite{ouali2020semi}. Only recently, attempts have been made focusing on consistency-based SSL for semantic segmentation~\cite{mittal2019semi}. In these methods, the `Teacher-Student' structure is often explored by creating a teacher model and a student model either explicitly or implicitly~\cite{zhang2021flexmatch,wang2021knowledge}. The core spirit is to impose consistency on the predictions between two models via an exponential moving average (EMA) of the student and force the unlabeled data to meet the smooth assumption in SSL.

However, such a learning paradigm suffers from three key problems. First, using the isomorphic CNN-based 
models show limited learning capability of consistency regularization (See Tab.~\ref{Table.2}). Previous works leveraged perturbations~\cite{french2019semi}, different initialization~\cite{ke2019dual} or different network structures~\cite{luo2021semi} to impose model diversity. However, as the two feature extractors are inevitably coupled, it is difficult for them to extract complementary features in the later stage of training. Moreover, existing SSL methods merely leverage the pixel-wise predictions from CNNs,
thus leading to a waste of rich inner knowledge in the feature space.
Lastly, the input for SSL is always the entire image, making it difficult to learn both global and local (\ie,~long-range) semantic information although operated by strong data augmentation.

It has been shown that Vision Transformer (ViT) can achieve comparable or even superior performance on image recognition tasks at a large scale~\cite{dosovitskiy2020image}.
Differing from CNNs consisting of convolutions (Convs), ViT's basic computational paradigm is multi-head self-attentions (MHSAs). Park \etal~\cite{park2021vision} show that MHSAs and Convs exhibit opposite behaviors. That is, there exist surprisingly clear differences in the features and internal structures of ViT and CNN~\cite{raghu2021vision}.

\textbf{Motivation:} Inspired by the success of ViT for visual recognition,
in this paper, we explore the potential of ViT and CNNs to tackle the above-mentioned problems for semi-supervised segmentation. However, bringing the ViT to SSL is challenging because: 
$a)$ the inner feature and output paradigm of ViT is heterogeneous from those of CNNs; 
$b)$ the high-performance of ViT needs the pre-training with hundreds of millions of annotated images using a large infrastructure~\cite{touvron2021training}; 
$c)$ in SSL, how to make Convs and MSAs learn with each other for the \textit{unlabeled} data from the feature and image level needs to be explored.  

To this end, we propose
a novel SSL method for semantic segmentation, called Transformer-CNN Cohort (\textbf{TCC}), by \textit{subtly incorporating the multi-level distillation to add consistency on the pixel-wise predictions and the heterogeneous feature space via pseudo labeling for the unlabeled data}. Specifically, as the ViT and CNN students have different input and inner feature flow forms, we notice that the feature maps extracted encode crucial complementary class-wise statistics~\cite{wang2020intra}. Therefore, we propose class-aware feature consistency distillation (\textbf{CFCD}) that first leverages the output of each student as the pseudo labels and generates pseudo prototype maps. Importantly, it is also the \textbf{\textit{first}} time that we explore pseudo labeling in SSL to facilitate feature distillation for the unlabeled data. 
The class-aware feature (CF) maps are computed by averaging the features on all pixels having the same pseudo labels. Class-aware feature variation knowledge is transformed via the CF maps between the cohort. Moreover, as the ViT student has more uniform representations for all layers, we propose Consistency-aware Cross Distillation (\textbf{CCD}) to distill knowledge based on the pixel-wise predictions from the cohort. As such, we can reduce the large amount of training data required for ViT and accommodate ViT to SSL tasks with high performance. We conduct extensive experiments with various settings on two benchmarks: PASCAL VOC 2012~\cite{pascal-voc-2012} and Cityscapes~\cite{Cordts2016Cityscapes}. The experimental results show that our TCC framework surpasses the existing SoTA methods by 4.15$\%$ under 1/16 label ratio on POASCAL VOC 2012 dataset and 1.03$\%$ under 1/16 partition protocols on the CityScapes dataset.

\textbf{Contributions}: In summary, the contributions of our paper are four-fold. (I) We propose the first SSL framework, with the transformer-CNN cohort, that imposes multi-level consistency on the pixel-wise predictions and the heterogeneous feature space. (II) We propose CFCD to distill the complementary class-wise feature knowledge via pseudo labeling for the \textit{unlabeled data}. Notably, we are also the \textit{first} to explore pseudo labeling for feature distillation in semi-supervised segmentation. (III) We propose CCD to distill the pixel-wise prediction knowledge to impose consistency for the students in the cohort. (IV) Our TCC framework achieves \textit{new} SoTA performance on both benchmarks. 

\section{Related work}

\noindent \textbf{Semi-supervised Semantic Segmentation.} 
Consistency regularization is widely applied for semi-supervised segmentation~\cite{sohn2020fixmatch,zheng2022uncertainty,chen2022uncertainty}. The key insight of this branch of approaches is that the predictions or intermediate features should be consistent across different semantic-preserving transformations on input or model of the same data. The image-level perturbation methods, \eg,~\cite{french2019semi,xie2023adversarial} randomly augment the input images while the feature-level perturbation methods, \eg,~\cite{ke2020gct} uses a multi-decoder strategy to augment the features. Moreover, CPS~\cite{chen2021semi} enforces consistency by using the pseudo segmentation maps with additional benefits like expanding the training data. 
Concurrently, \cite{luo2021semi} proposes SSL-based method for medical imaging using ViT's and CNN's predictions. 
Differently, we propose the first SSL method for semantic segmentation by exploring the potential of ViT and CNN and subtly incorporating the multi-level consistency distillation on the pixel-wise predictions and the heterogeneous feature space via pseudo labeling for the unlabeled data.

\noindent \textbf{Vision Transformer.}
Transformer was proposed by Vaswani et al.~\cite{vaswani2017attention} to solve the machine translation tasks. 
Several works have applied ViT to high-level vision tasks, \eg, object detection~\cite{beal2020toward,dai2021up,sun2021rethinking,zhu2020deformable} and semantic segmentation~\cite{zheng2021rethinking,wu2020visual,chen2021pre,wang2021end}. Recently, PVT~\cite{wang2021pyramid} introduces the pyramid structure into ViT to generate multi-scale features for dense prediction tasks. 
ViT has been continually improved and achieved better performance on large-scale datasets~\cite{touvron2021training}.
However, ViT does not generalize well in case of insufficient data~\cite{dosovitskiy2020image,zhu2023good,zheng2023distilling}. Therefore, pre-training on a curated data is required for training an efficient ViT model. To tackle this issue, Bao et al.~\cite{bao2021beit} introduce a masked image modeling approach to the pre-trained ViT while Touvron et al.~\cite{touvron2021training} explore a hard distillation method.
We explore the potential of ViT and incorporate it with CNNs as a cohort for semi-supervised semantic segmentation.
Our TCC framework subtly imposes the multi-level consistency distillation on the pixel-wise predictions and the heterogeneous feature space via pseudo labeling for the unlabeled data. 

\noindent \textbf{Knowledge Distillation (KD)}
aims to build a smaller (student) model with the softmax labels of a larger (teacher) model~\cite{hinton2015distilling,zheng2023distilling}. 
There are several paradigms in KD, including soft distillation~\cite{hinton2015distilling}, hard-label distillation, and label smoothing~\cite{szegedy2016rethinking}. 
Some works~\cite{passalis2018learning,peng2019correlation,tian2019contrastive} explore the structural information within the feature space to learn more generic representation. DeiT~\cite{touvron2021training} first introduces a KD method specific to ViT aiming to distill the token. It shows that using the CNNs as teachers achieves better performance than using the ViT models mainly because of the inductive bias brought by convolution (Convs). 
Recently, Raghu \etal.~\cite{raghu2021vision} analyzes the representation structure of ViT and CNN on the visual recognition tasks and finds striking differences between the two models. That is, Convs and MSAs are two ways of extracting features, making CNNs and ViT sensitive to different features. Applying KD to CNNs and ViT cohort in semi-supervised segmentation is challenging as the inputs and learning capability for both students are different; we thus propose CFCD that leverages the feature maps via pseudo prototype from each student and transfers the complementary class-wise information.

\begin{figure*}[t!]
\centering
\includegraphics[width=0.8\linewidth]{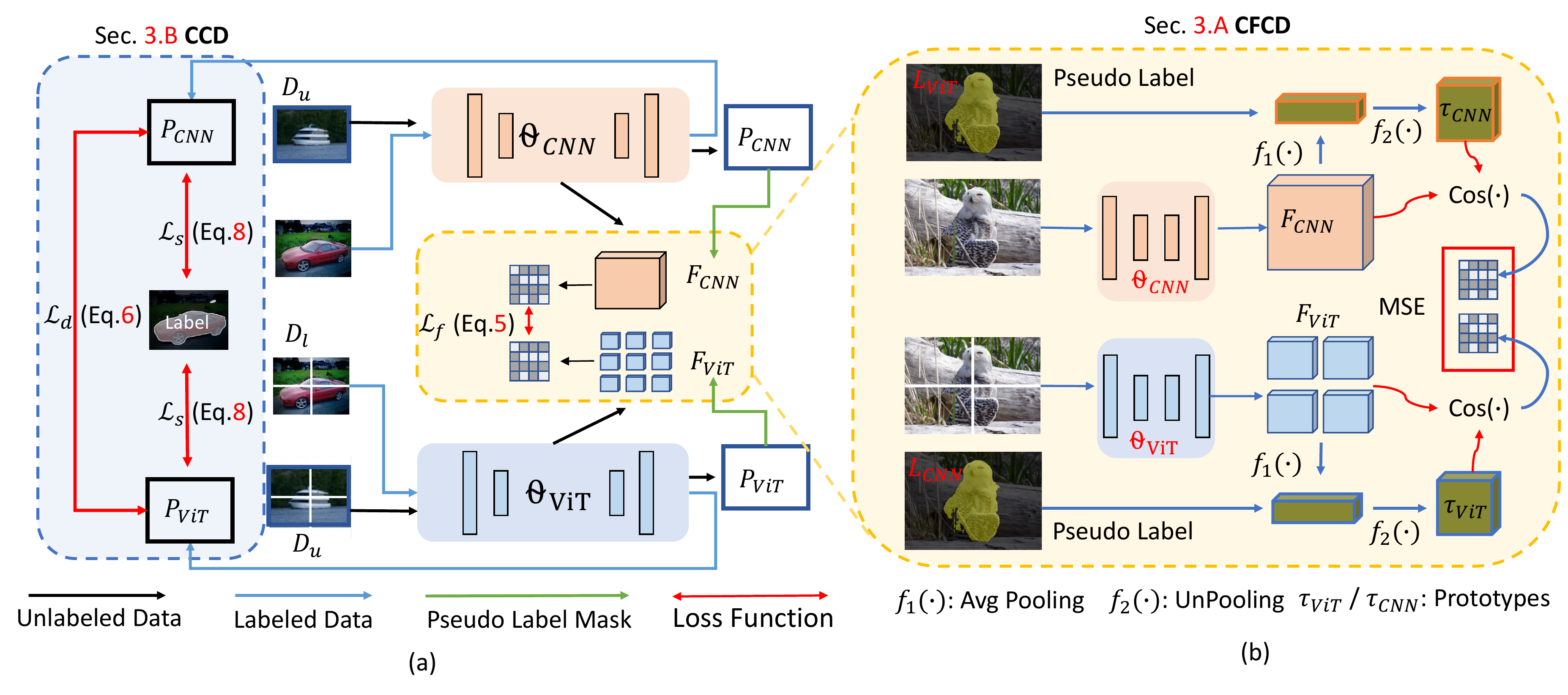}
\vspace{-10pt}
\caption{(a) The proposed TCC framework comprises two students: the ViT  $f(X; \theta_{ViT})$ and the CNN student $f(X;\theta_{CNN})$. The dashed lines indicate the fully supervised learning of the cohort with limited labeled data. TCC incorporates multi-level consistency distillation for the pixel-wise predictions (CCD) and class-aware feature consistency distillation (CFCD) for the unlabeled data. (b) The proposed CFCD involves two networks with different class-wise feature variations, characterized by the similarity between the feature on each pixel and its corresponding class-wise prototype (dashed lines). The higher similarity between the prototypes results in lower variation. The two networks, $f(\theta_{ViT})$ and $f(\theta_{CNN})$, learn from each other accordingly.
} 
\label{overall_TCC}
\vspace{-10pt}
\end{figure*}

\section{The Proposed Approach}
An overview of the proposed TCC framework is shown in Fig.~\ref{overall_TCC} (a). Given a labeled set $D_l$ of $N$ labeled images and a set $D_u$ of $M$ unlabeled images, we propose
the first yet novel SSL method for semantic segmentation, called Transformer-CNN cohort (TCC), by subtly incorporating the multi-level distillation to add consistency on the pixel-wise predictions and the heterogeneous feature space via pseudo labeling for the unlabeled data. 
Our TCC framework consists of two students: the ViT student $f(X; \theta_{ViT})$ and the CNN student $f(X;\theta_{CNN})$. 
That is, given the input images $X$, we aim to attain the segmentation confidence maps ($P_{CNN}$ and $P_{ViT}$), high-level features ($F_{CNN}$ and $F_{ViT}$) and pseudo labels($L_{CNN}$ and $L_{ViT}$) from the cohort  $f(X; \theta_{ViT})$ and $f(X;\theta_{CNN})$, which can be formulated as: 
\begin{align}
\setlength{\abovedisplayskip}{3pt}
\setlength{\belowdisplayskip}{3pt}
    L_{CNN},P_{CNN},F_{CNN} &= f(X;\theta_{CNN});  \\
    L_{ViT},P_{ViT},F_{ViT} &= f(X;\theta_{ViT}).   
\end{align}

Our key ideas are three folds. Firstly, as $f(X; \theta_{ViT})$ and $f(X;\theta_{CNN})$ have different inputs and inner feature forms, the  extracted feature maps encode crucial complementary class-wise statistics. Therefore, we propose class-aware feature consistency distillation (CFCD) that first leverages the output of each student as the pseudo labels and generates feature prototype maps.
Note that it is the \textit{first time we explore pseudo labelling in SSL to facilitate feature distillation for the unlabeled data}. The class-aware features maps are computed by averaging the features on all pixels having the same pseudo labels. Feature variation knowledge is transformed via the cohort's class-aware feature (CF) maps. Secondly, as $f(X; \theta_{ViT})$ possesses more uniform representations for all layers~\cite{raghu2021vision}, we propose Consistency-aware Cross Distillation (CCD) that distills the pixel-wise prediction information bidirectionally based on the heterogeneous students.
Lastly, similar to other SSL methods, \eg,~\cite{sohn2020fixmatch}, supervised training is applied to both $(X; \theta_{ViT})$ and $f(X;\theta_{CNN})$ for the limited labeled data.
We now describe these components in detail.
\subsection{Class-aware Feature Consistency Distillation}
\label{3.2}
\noindent \textbf{Pseudo Prototype}
Theoretically, high-dimensional feature representations obtained by two different models should be distinct explicitly but share implicit commonalities~\cite{wang2020intra}. For all pixels of the same class in the corresponding class-wise label maps, their mapping center in feature space is referred to as the class-wise prototype. \textit{This prototype is based on the condition that all pixels belonging to the same class should coincide in feature mapping space}. However, the feature maps of all pixels of the same class may not fall on the prototype completely; therefore, we estimate the class-aware feature variation to measure the similarity between the mapping of each pixel and the prototype. The class-aware feature variation for pixels can be obtained from the predictions and the inner features of $f(X; \theta_{ViT})$ and $f(X; \theta_{CNN})$. As features reflects how students in the cohort understand the input, it is crucial that reliable prototype calculation is guaranteed.
However, for the unlabeled data, there are no prior labels for computing the prototype in SSL; thus, we leverage the pseudo labels predicted from one student as the source prototype for the other. That is, the CF map obtained from $f(X; \theta_{ViT})$ is taken as the standard prototype for student $f(X; \theta_{CNN})$ in the feature space, and vice versa. As such, $f(X; \theta_{ViT})$ and $f(X; \theta_{CNN})$ can be better correlated to each other despite of the difference of computing paradigms (MHSAs for $f(X; \theta_{ViT})$ and Convs for $f(X; \theta_{CNN})$). 

As shown in Fig.~\ref{overall_TCC} (b), to impose class-aware feature consistency to both students in the cohort, pseudo labels are down-sampled with nearest neighbour interpolation to match the spatial size of the high-dimensional features. Then, average pooling is operated on the masked features, corresponding to pixels with the same label for each class to get the class-wise pseudo prototype.
Finally, we perform average pooling on the masked region of each pseudo prototype to ensure each position stores the corresponding high-dimensional feature of the class-wise prototype. Overall, the prototype can be formulated as:
\begin{align}
\setlength{\abovedisplayskip}{3pt}
\setlength{\belowdisplayskip}{3pt}
    &\textit{$\mathcal{T}$}_{ViT} = \textit{f}_2(\textit{f}_1(\textit{Mask}(F_{ViT},L_{CNN}))); \\
    &\textit{$\mathcal{T}$}_{CNN} = \textit{f}_2(\textit{f}_1(\textit{Mask}(F_{CNN},L_{ViT}))),
\end{align}
where $\mathcal{T}_{ViT}$ and $\mathcal{T}_{CNN}$ denote the pseudo prototypes for the students $f(X; \theta_{ViT})$ and $f(X; \theta_{CNN})$, respectively; $\textit{f}_1(\cdot)$ is the average pooling; $\textit{f}_2$($\cdot$) is the unpooling operation; $F_{ViT}$ and $F_{CNN}$ are the high-dimensional features masked by pseudo label maps $L_{CNN}$ and $L_{ViT}$, respectively. 
As such, we can obtain CF maps via calculating the similarity,~\eg, cosine similarity, between $\mathcal{T}_{CNN}$ and $\mathcal{T}_{CNN}$ for each student, as shown in Fig.~\ref{overall_TCC} (b). More details of CF map calculation will be described in the following section. 

\noindent \textbf{Feature Distillation via Pseudo Prototype}
The CNN kernel of $f(X; \theta_{CNN})$ inspects adjacent pixels and gradually expands its receptive field to a more significant portion of an image, producing features with high locality. By contrast, $f(X; \theta_{ViT})$ manipulates patch-level image at every stage, having a global vision even at the beginning. Though the two models achieve similar performance after the training, their learning process is distinctive. Moreover, the feature maps extracted from them encode vital complementary class-wise statistics.
Intuitively, we find that transferring the feature-level knowledge can complement the drawbacks of each model and thus yield better results.
Inspired by~\cite{wang2020intra}, we apply consistency regularization on the pseudo feature variation between $f(X; \theta_{ViT})$ and $f(X;\theta_{CNN})$, as depicted in Fig.~\ref{overall_TCC} (b).
Especially, we extract the high-dimensional features output from the last stage of $f(X; \theta_{ViT})$ and $f(X;\theta_{CNN})$ to calculate their corresponding class-wise prototype. The prototype $\mathcal{T}$ at pixel $p$ of class $c$ is computed by averaging the features on all pixels having class label $c$, given by Eq.~\ref{Eq.3}:

\begin{equation}
\setlength{\abovedisplayskip}{-6pt}
\setlength{\belowdisplayskip}{3pt}
   \mathcal{T}(i) = \frac{1}{|S_c|} \sum_{i\in S_c}f(i)
    \label{Eq.3}
\end{equation}
\begin{equation}
\setlength{\abovedisplayskip}{3pt}
\setlength{\belowdisplayskip}{3pt}
    M(i) = \textit{Cos}(f(i),\mathcal{T}(i)); 
    \label{Eq.4}
\end{equation}

where $f(i)$ denotes the feature on pixel $i$, $S_c$ is the set of pixels having the label $c$, $|S_c|$ stands for the size of the set $S_c$, $\mathcal{T}(i)$ is the class-wise feature variation map from its cohort, and $M(i)$ denotes the value of class-wise feature variation map at pixel $i$. In Eq.~\ref{Eq.4}, due to the intrinsic difference (\eg, magnitude, deviation) in the feature maps between $f(X; \theta_{ViT})$ and $f(X;\theta_{CNN})$, we adopt the cosine similarity \textit{Cos}($\cdot$) to better formulate the relative distribution for each class.
Finally, the CFCD loss $\mathcal{L}_f$ minimizes the distance between feature variation maps of the students $f(X; \theta_{ViT})$ and $f(X;\theta_{CNN})$. Specifically, we employ the Mean Squared Error (MSE) loss as follows:
\begin{equation}
\setlength{\abovedisplayskip}{3pt}
\setlength{\belowdisplayskip}{3pt}
     \mathcal{L}_f = \frac{1}{N} \sum_{p\in\Omega}(M_{CNN}(i) - M_{ViT}(i))^2,
\end{equation}
where $N$ is the total numbers of pixels, $\Omega$ denotes the image, $M_{CNN}(p)$ and $M_{ViT}(p)$ represent the corresponding pseudo feature variation map of the CNN and ViT students.

\subsection{Consistency-aware Cross Distillation}
\label{3.3}
Though $f(\theta_{ViT})$ and $f(\theta_{CNN})$ share different learning capacities for the unlabeled data $D_u$, their predictions should be consistent according to the \textit{smoothness assumption} where samples in the same cluster are expected to have the same labels. In particular, instead of using the Exponential Moving Average~\cite{tarvainen2017mean} to update the predictions, we find that measuring the cross-model discrepancy between $f(\theta_{ViT})$ and $f(\theta_{CNN})$ helps improve each student's representations.
Accordingly, we propose Consistency-aware Cross Distillation (CCD) to enforce consistency between the outputs of the cohort to extract additional information for the unlabeled data $D_u$, as shown in Fig.~\ref{Figure.1} (a). CCD is bidirectional: one is from $f(\theta_{CNN})$ to $f(\theta_{ViT})$ and the other one is from $f(\theta_{ViT})$ to $f(\theta_{CNN})$. That is, we use the logits output $P_{CNN}$ from the CNN student $f(\theta_{CNN})$ to supervise the logits output $P_{ViT}$ of the ViT student $f(\theta_{ViT})$, and vice versa. The CCD loss on the unlabeled data $D_u$ is:
\begin{equation}
\setlength{\abovedisplayskip}{3pt}
\setlength{\belowdisplayskip}{3pt}
    \mathcal{L}_{d}^{u} = \frac{1}{|D_l|}(\mathcal{L}_{kl}(P_{CNN},P_{ViT}) + \mathcal{L}_{kl}(P_{ViT},P_{CNN})),  
\end{equation}
where the $\mathcal{L}_{kl}$ is the standard KL divergence. The CCD loss $\mathcal{L}_{d}^{l}$ on the labeled data $D_l$ can be defined in the same manner. The total CCD loss is the combination of losses on both the labeled $D_l$ and unlabeled data $D_u$: $\mathcal{L}_d$ = $\mathcal{L}_{d}^{l}$ + $\mathcal{L}_{d}^{u}$.

\subsection{Optimization}
\label{3.4}
The training objective contains three losses as follows:
\begin{equation}
\setlength{\abovedisplayskip}{3pt}
\setlength{\belowdisplayskip}{3pt}
\label{eq:total}
    \mathcal{L} = \mathcal{L}_s + g(t)\cdot (\mathcal{L}_d + \lambda \mathcal{L}_f),
\end{equation}
where the $\mathcal{L}_s$ is supervised loss, the $\mathcal{L}_d$ refers to the prediction-level CCD loss and the $\mathcal{L}_f$ is the CFCD loss that measures the class-aware feature variation consistency between $f(\theta_{ViT})$ and $f(\theta_{CNN})$. The $g(t)$ is a consistency ramp-up function following \cite{laine2016temporal}, and $\lambda$ is a fixed constant. The supervision loss $\mathcal{L}_s$ is formulated as:
\begin{equation}
    \mathcal{L}_s = \frac{1}{|D_l|}\sum_{X\in D_l }(\mathcal{L}_{dice}(P_{CNN},Y_{CNN}) + \mathcal{L}_{dice}(P_{ViT},Y_{ViT}))
\end{equation}
where $\mathcal{L}_{dice}$ indicates the Dice coefficient loss function and $Y$ are the ground truth (GT) labels.

\section{Experiments and Evaluation}
\begin{table}[]
    \setlength{\tabcolsep}{4mm}
    \centering
        \caption{\textbf{Comparison with state-of-the-arts} on the PASCAL VOC.} 
    \resizebox{\linewidth}{!}{
    \begin{tabular}{cccccc}
    \toprule
    \multirow{2}{*}{Method} & \multirow{2}{*}{Backbone}            & \multicolumn{4}{c}{Label Rate}                \\ \cmidrule{3-6} 
                            &                                      & 1/16 & 1/8 & 1/4 & 1/2 \\ \midrule
    \multirow{2}{*}{CCT~\cite{ouali2020semi}}    & ResNet-50                            & 65.22     & 70.87     & 73.43     & 74.75     \\ 
                            & ResNet-101                           & 67.94     & 73.00     & 76.17     & 77.56     \\ \midrule
    \multirow{2}{*}{CPS~\cite{chen2021semi}}    & ResNet-50                            & 68.21     & 73.20     & 74.24     & 75.91     \\ 
                            & ResNet-101                           & 72.18     & 75.83     & 77.55     & 78.64     \\ \midrule
    \multirow{2}{*}{$n$-CPS~\cite{ncps}}    & ResNet-50                            & 68.36     & 73.45     & 75.75     & 77.00     \\ 
                            & ResNet-101                           & 73.51     & 76.46     & 78.59     & 79.90     \\ \midrule
    
    AEL~\cite{AEL}                     & ResNet-101                           & 77.20     & 77.57     & 78.06     & 80.29     \\ \midrule
    \multirow{2}{*}{ST++~\cite{yang2022st++}}   & ResNet-50                            & 73.20     & 75.50     & 76.00     & -         \\ 
                             & ResNet-101                           & 74.70     & 77.90     & 77.90     & -         \\ \midrule
    $U^2PL$~\cite{U2PL}                    & ResNet-101                           & 77.21     & 79.01     & 79.30     & 80.50     \\ \midrule
     \multirow{2}{*}{ELN~\cite{ELN}}    & ResNet-50                            & 70.52     & 73.20     & 74.63     & -         \\ 
                             & ResNet-101                           & 72.52     & 75.10     & 76.58     & -         \\ \midrule
     PAMT~\cite{liu2022perturbed}   
    & ResNet-101                           & 75.50     & 78.20     & 78.72     & 79.76         \\ \midrule
    CSS~\cite{wang2023space}   
    & ResNet-101                           & 78.73     & 79.54     & 80.82     & 81.06         \\ \midrule
    \multirow{4}{*}{Ours}    & TCC-S   & 79.16      & 80.28     & 82.32      & 82.52      \\
    & w/ Cutmix   &\textbf{81.35}      &\textbf{83.05}      &\textbf{83.55}      & \textbf{84.04}     \\
    \cmidrule{2-6}
                             & TCC-B                               & 80.17     & 81.17     & 82.42     & 82.80     \\ 
                            & \multicolumn{1}{l}{w/ Cutmix} & \textbf{81.36}     & \textbf{83.42}     & \textbf{84.27}     & \textbf{84.29}     \\ \bottomrule
    \end{tabular}}
    \label{tab:voc}
    \vspace{-8pt}
    \end{table}
    
    \begin{table}[t!]
    \setlength{\tabcolsep}{4mm}
    \centering
        \caption{\textbf{Comparison with the state-of-the-arts} on the Cityscapes
       val set under different partition protocols. All the methods are based on DeepLabv3+. (TCC-B: TCC-Base; w/ Cutmix: with cutmix augmentation)} %
    \resizebox{\linewidth}{!}{
    \begin{tabular}{cccccc}
    \toprule
    \multirow{2}{*}{Method} & \multirow{2}{*}{Backbone}            & \multicolumn{4}{c}{Label Rate}                \\ \cmidrule{3-6} 
                            &                                      & 1/16 & 1/8 & 1/4 & 1/2 \\ \midrule
    MT~\cite{tarvainen2017mean}
                            & ResNet-101                           & 68.08     & 73.71     & 76.53     & 78.59     \\ 
    CCT~\cite{ouali2020semi}
                            & ResNet-101                           & 69.64     & 74.48     & 76.35     & 78.29     \\ 
    CPS~\cite{chen2021semi}
                            & ResNet-101                           & 70.50     & 75.71     & 77.41     & 80.08     \\ 
    3-CPS~\cite{ncps}                     & ResNet-101                           & 75.86     & 77.99     & 78.95     & 80.26     \\ 
    AEL~\cite{AEL}                     & ResNet-101                           & 75.83     & 77.90     & 79.01     & 80.28     \\
    $U^2$PL~\cite{U2PL}                    & ResNet-101                           & 70.30     & 74.37     & 76.47     & 79.05     \\ \midrule
    \multirow{2}{*}{Ours}   
                            & TCC-B                               & 75.79     & 77.53     & 78.47     & 80.83     \\ 
                            & \multicolumn{1}{l}{w/ CutMix} & \textbf{76.89}     & \textbf{78.52}     & \textbf{80.04}     & \textbf{80.93}     \\ \bottomrule
    \end{tabular}}
    \label{Table.2}
    \end{table}

\textbf{Datasets}. \textbf{Pascal VOC} contains 20 foreground object classes plus an extra background class. The standard training, validation, and test sets consist of 1464, 1449, and 1456 images, respectively. We adopt the augmented set from \cite{hariharan2011semantic} which contains 10582 images as our full training set.
\textbf{Cityscapes} contains a diverse set of video sequences recorded in street scenes from 50 cities, with high-quality pixel-level annotations. The official split has 2975 images for training, 500 for validation, and 1525 for testing. Each image is finely annotated with pixel-level labels of 19 classes. We divide the whole training set into two groups by randomly sampling 1/2, 1/4, 1/8, and 1/16 of the entire set as labeled images and the rest as unlabeled images. For a fair comparison, images in each set are the same as CPS~\cite{chen2021semi}.

\noindent \textbf{Evaluation and comparison}. We leverage the mean Intersection-over-Union (mIOU) as an evaluation metric. Our trained models are evaluated on PASCAL VOC 2012 validation set (1456 images) and the \emph{Cityscapes} validation set (500 images) via testing at a single scale, respectively. We report the mIOU of the ViT-based model in the cohort.  For comparison, in all tables, \textit{\textbf{ResNet50, PVT-M, and ConvNext-S mean that two students in our TCC are based on the same backbone. Also, all the compared methods are implemented with dual students}}.
\begin{figure*}[t!]
     \centering
     \includegraphics[width=0.80\textwidth, height=3.5cm]{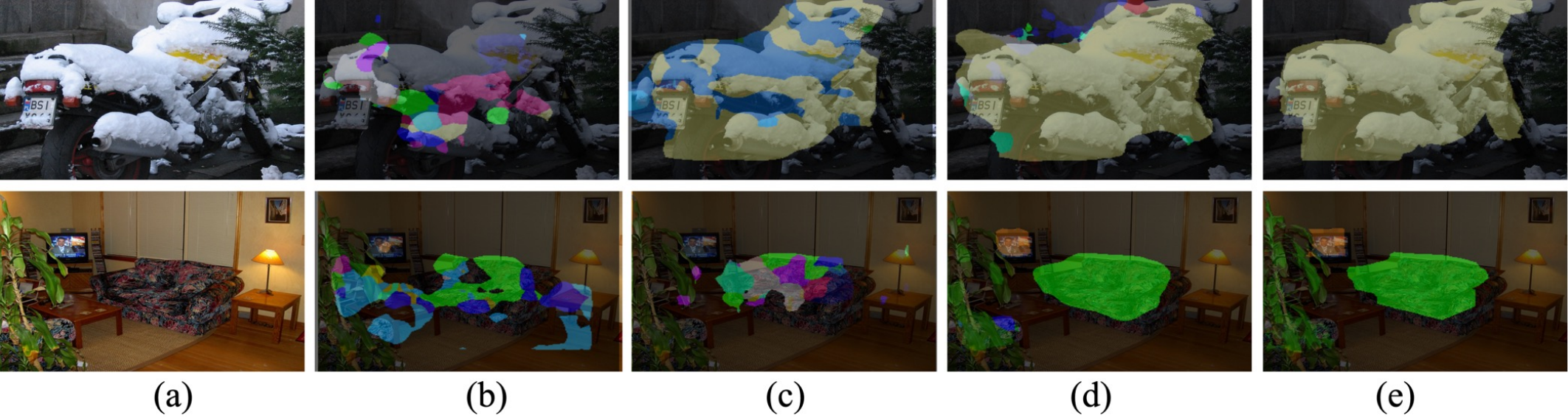}
     \vspace{-6pt}
     \caption{\textbf{Example results from PASCAL VOC 2012}. (a) input, (b) CCT~\cite{ouali2020semi}, (c) CPS~\cite{chen2021semi}, (d) ours (TCC), and (e) ground truth. All the approaches are based on DeepLabv3+.
     }
     \label{Figure.4}
     \vspace{-8pt}
 \end{figure*}
\noindent \textbf{Implementation details}. We implement our TCC framework using Pytorch. We initialize the weights of two backbones, \ie, the students  $f(\theta_{ViT})$ and $f(\theta_{CNN})$, with the weights pre-trained on ImageNet $1K$ and the weights of two segmentation heads (of DeepLabv3+~\cite{chen2017deeplab}) randomly. We use the AdamW optimizer and train the TCC framework for 30000 iterations with a total batch size of 16 for the PASCAL VOC dataset and 80000 iterations with a total batch size of 4 for Cityscapes. We employ the poly learning rate policy where the initial learning rate $\alpha$ is multiplied by $(1 - \frac{iteration}{max iterations})^{0.9} $, and $\lambda$ is simply set to 1 for both datasets.
For fair comparisons, two types of settings in our TCC are settled to be compared with ResNet-50 and ResNet-101, \ie, Cohort-Small (TCC-S) and Cohort-Base (TCC-B). Specifically, \textit{ResNet-50 and PVT-S are the two backbones in TCC-S, while ConvNeXt-S and PVT-M in TCC-B}. We also report the results of ResNet, PVT, and ConvNext-based implementations of TCC in Tab.~\ref{differbackbone}.
\subsection{Comparison with the SoTA methods}
We compare our method with the SoTA semi-supervised methods including Mean-Teacher(MT)~\cite{tarvainen2017mean}, Cross Consistency Training(CCT)~\cite{ouali2020semi}, Cross Pseudo Supervision(CPS)~\cite{chen2021semi}, AEL~\cite{AEL}, ST++~\cite{yang2022st++}, $U^2$PL~\cite{U2PL} and ELN~\cite{ELN} under different label ratios. 

Our method even outperforms $n$-CPS\cite{ncps}, a 3-model based method, by 7.85$\%$, 6.96$\%$, 5.68$\%$, and 4.90$\%$, respectively, under the same labeled ratios on \emph{Pascal VOC} dataset. The greatest improvement at the 1/16 label ratio indicates that the combination of CNN and ViT students can facilitate each other when learning unlabeled data and generalize well on labeled data, which is consistent with our assumption. 

\noindent \textbf{Detailed results by datasets.}
\textit{PASCAL VOC 2012}: Tab.~\ref{tab:voc} shows the comparison results. On all label ratios (1/2, 1/4, 1/8, and 1/16), our TCC approach (w/o CutMix) consistently outperforms the other methods. And our TCC approach (w/ CutMix) achieves the best performance and sets new SoTA under all label ratios. With lower labeled ratios, our approach (w/ CutMix) outperforms the $U^2$PL by $4.15\%$ and $4.41\%$, respectively, under 1/16 and 1/8 label ratios. 
\textbf{\textit{This confirms that using two heterogeneous models, \ie, ViT and CNN, in the consistency regularization approaches achieves better performance, especially with less annotated data}}. 
Fig.~\ref{Figure.4} shows the qualitative results. CCT struggles to catch the main objects from inputs and wrongly classifies many regions of interest into the background (black). It renders a colorful mask over a single object and is especially devastating if some natural camouflage exists, \eg, the snow-covered motorbike. CPS performs relatively better than CCT. It can roughly outline the boundary of objects of interest but the inner pixel-wise segmentation results are still heterogeneous for a single object. In contrast, our method achieves much neater and cleaner segmentation results (4th column), which is much closer to the GT label maps (5th column).

\begin{figure*}[]
     \centering
     \includegraphics[width=0.8\textwidth, , height=4cm]{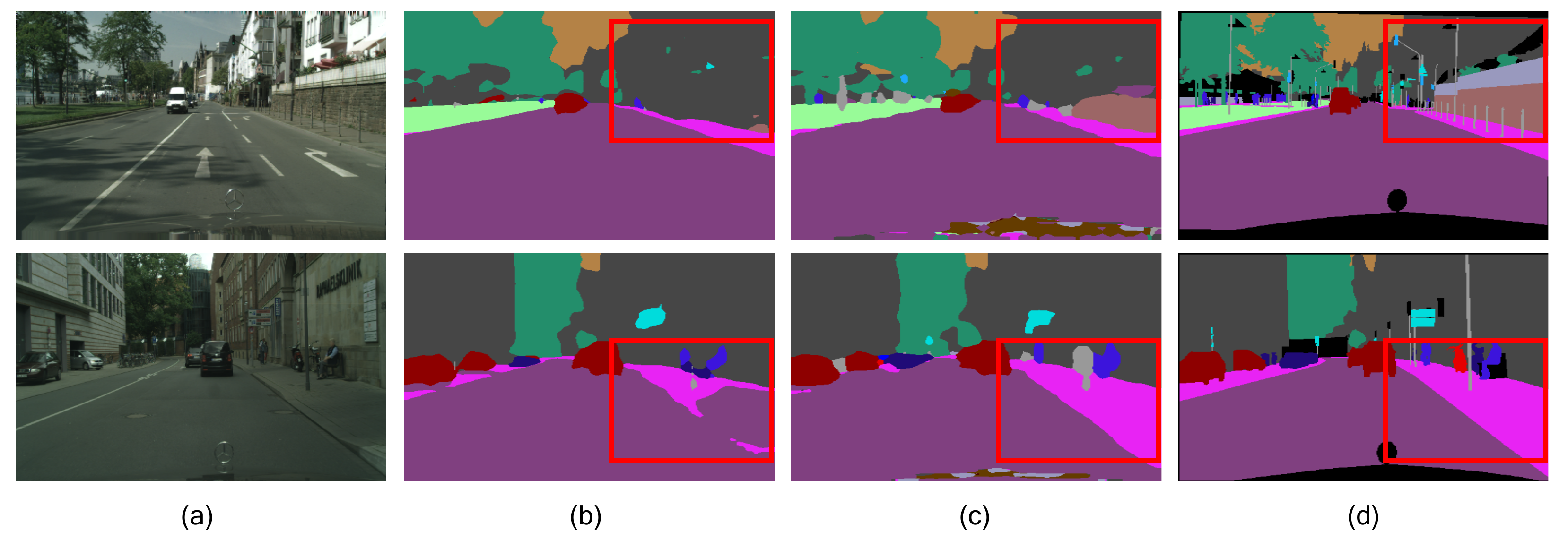}
     \vspace{-6pt}
     \caption{\textbf{Example results from Cityscapes}. (a) input, (b) CPS~\cite{chen2021semi}, (c) ours, and (d) GT. All methods are based on DeepLabv3+.}
     \vspace{-6pt}
     \label{fig:city_results}
\end{figure*}

\textit{Cityscapes $val$ set}: Tab.~\ref{Table.2} shows the quantitative results where our TCC approach consistently outperforms the SoTA methods.
The improvements of mIOU of our method (w/o CutMix augmentation) over the 2-model baseline method (AEL) are 1.06$\%$, 0.62$\%$, 1.03$\%$, 0.65$\%$ under label ratio of 1/16, 1/8, 1/4, and 1/2, respectively. The qualitative results are shown in Fig.~\ref{fig:city_results}.

\subsection{Ablation study and Analysis}

\begin{table}[]
\setlength{\tabcolsep}{4mm}
\centering
\caption{Different loss combinations on PASCAL VOC.}
\resizebox{\linewidth}{!}{
\begin{tabular}{ccccccc}
\toprule
\multicolumn{3}{c}{Losses} & \multicolumn{4}{c}{PASCAL VOC 2012}           \\ \midrule
$\mathcal{L}_s$ &$\mathcal{L}_d$ &$\mathcal{L}_f$      & 1/16 & 1/8 & 1/4 & 1/2 \\ \midrule
\checkmark &         &        &74.65 & 76.22     & 76.71      & 77.01          \\ \midrule
\checkmark& \checkmark &        &77.90 & 80.46& 81.60  & 81.95          \\ \midrule
\checkmark& \checkmark& \checkmark &\textbf{80.17}  &\textbf{81.17}  &\textbf{82.42} &\textbf{82.80}           \\ \bottomrule
\end{tabular}}
\label{Table.3} 
\end{table}
    
\begin{table}[]
    \setlength{\tabcolsep}{4mm}
    \centering
    \caption{Comparison of our TCC framework trained form scratch (without pre-trained models) and previous SoTA semi-supervised segmentation methods.}
    \resizebox{\linewidth}{!}{
\begin{tabular}{lcccc}
\toprule
Method                   & 1/16 & 1/8 & 1/4 & 1/2 \\ \midrule
MT\cite{tarvainen2017mean} & 66.77     & 70.78     & 73.22     & 75.41     \\ 
CCT\cite{ouali2020semi}  & 65.22     & 70.87     & 73.43     & 74.75     \\
CPS\cite{chen2021semi}   & 68.21     & 73.20     & 74.24     & 75.91     \\
CPS*\cite{chen2021semi}  & 74.18     & 74.19     & 74.76     & 78.37     \\ 
3-CPS-mc\cite{ncps}&68.36      & 73.45     &75.75      &77.00     \\
3-CPS-mc\cite{ncps}&72.03      & 74.18     &75.85      &76.65     \\ \midrule
Ours(w/ pre-train)   &\textbf{80.17}     & \textbf{81.17}     &\textbf{82.42}     &\textbf{82.80}     \\ 
Ours(w/o pre-train)   & \textbf{78.61}  & \textbf{79.43}     & \textbf{80.09}   & \textbf{80.14}       \\ 
\bottomrule
\end{tabular}}
\label{tab:scratch}
\end{table}
    
\noindent \textbf{Improving Few-Supervision}
\begin{table}[]
    \setlength{\tabcolsep}{4.5mm}
    \centering
    \caption{Comparison for few-supervision on PASCAL VOC dataset.}
    \resizebox{\linewidth}{!}{
\begin{tabular}{lcccc}
\toprule
\multirow{2}{*}{Method} & \multicolumn{4}{c}{labeled samples} \\ \cmidrule{2-5} 
                        & 1/2 & 1/4 & 1/8 & 1/16      \\ \midrule
AdvSemSeg\cite{hung2018adversarial}& 65.27    & 59.97    & 47.58   & 39.69   \\ 
CCT\cite{ouali2020semi} & 62.10    & 58.80    & 47.60   & 33.10   \\ 
GCT\cite{ke2020gct}     & 70.67    & 64.71    & 54.98   & 46.04   \\ 
VAT\cite{miyato2018virtual}& 63.34    & 56.88    & 49.35   & 36.92   \\ 
CutMix-Seg\cite{french2019semi}& 69.84    & 68.36    & 63.20   & 55.58   \\ 
PseudoSeg\cite{zou2020pseudoseg}& 72.41    & 69.14    & 65.50   & 57.60   \\ 
CPS+CutMix\cite{chen2021semi}    & 75.88    & 71.71   & 67.42   &64.07    \\\midrule
Ours   & 77.92    & 74.25    & 72.99   & 72.01        \\ 
\textbf{Ours w/ CutMix}  &\textbf{82.80}    & \textbf{76.92}    & \textbf{74.61}   &\textbf{73.39}         \\ \bottomrule
\end{tabular}}
\label{tab:few}
\end{table}
Since our TCC framework outperforms the SoTA methods with less labeled data, as mentioned above, we study the performance of TCC on the PASCAL VOC 2012 with few labels by following the same partition in PseudoSeg~\cite{zou2020pseudoseg} which randomly samples $1/2$, $1/4$, $1/8$, and $1/16$ of images in the standard training set (1464 images) to construct the labeled set. The remaining images are used as an unlabeled set. We report the results of our approach (w/ and w/o CutMix augmentation). The results are listed in Tab.~\ref{tab:few}. The improvements of mIOU of our model (w/ CutMix augmentation) over CPS~\cite{chen2021semi} are 6.92$\%$, 5.21$\%$, 7.19$\%$ and 9.32$\%$, respectively, under 1/16, 1/8, 1/4, and 1/2 label ratios.
Our method achieves the best results and is superior to CPS again on the few labeled case. This validates the fact that our TCC subtly incorporates the multi-level distillation to add consistency on the pixel-wise predictions and the heterogeneous feature space via pseudo labeling for the unlabeled data.

\noindent \textbf{Train from scratch}
\textit{We also train our TCC framework without pre-trained models.} Tab.~\ref{tab:scratch} demonstrates that, even without ImageNet 1K pre-trained models, our TCC framework outperforms the previous SoTA method with pre-trained models by a large margin.

\noindent \textbf{Loss functions.} We conduct ablation experiments on the PASCAL VOC 2012 to analyze the impact of the CCD loss $\mathcal{L}_d$ and CFCD loss $\mathcal{L}_f$ in our TCC framework.
In Tab.~\ref{Table.3}, different combinations of losses are applied. We can see that TCC framework leverages the unlabeled data well with an average improvement of 5.49$\%$ over fully supervised by labeled data. Both $\mathcal{L}_d$ and $\mathcal{L}_f$ contribute positively to the validation mIOU in every label rate. 
We notice that the increases brought by  $\mathcal{L}_f$ decline when the number of unlabeled data decreases. It is also intuitive since CF maps in $\mathcal{L}_f$ is inferred by unlabeled data. Reduction of the hampers models fitting the actual class-wise feature variance distribution, thus lowering the improvement downward.

\noindent \textbf{Difference with CPS~\cite{chen2021semi}}
Our TCC shares a quite different spirit with CPS in that \textbf{ 1)} ours imposes \textit{multi-level} consistency on the pixel-wise predictions and heterogeneous features, but CPS merely utilizes the prediction-level cross-pseudo supervision which neglects the feature-level knowledge; \textbf{2)} we are the \textbf{\textit{first}} to explore pseudo labeling for feature distillation; \textbf{3)} we are the \textbf{\textit{first}} to study how heterogeneous models (CNNs and ViTs) benefit SSL performance. 
Note that CPS only utilizes the same backbone.

\begin{figure}[t]
\centering
\includegraphics[width=0.75\linewidth]{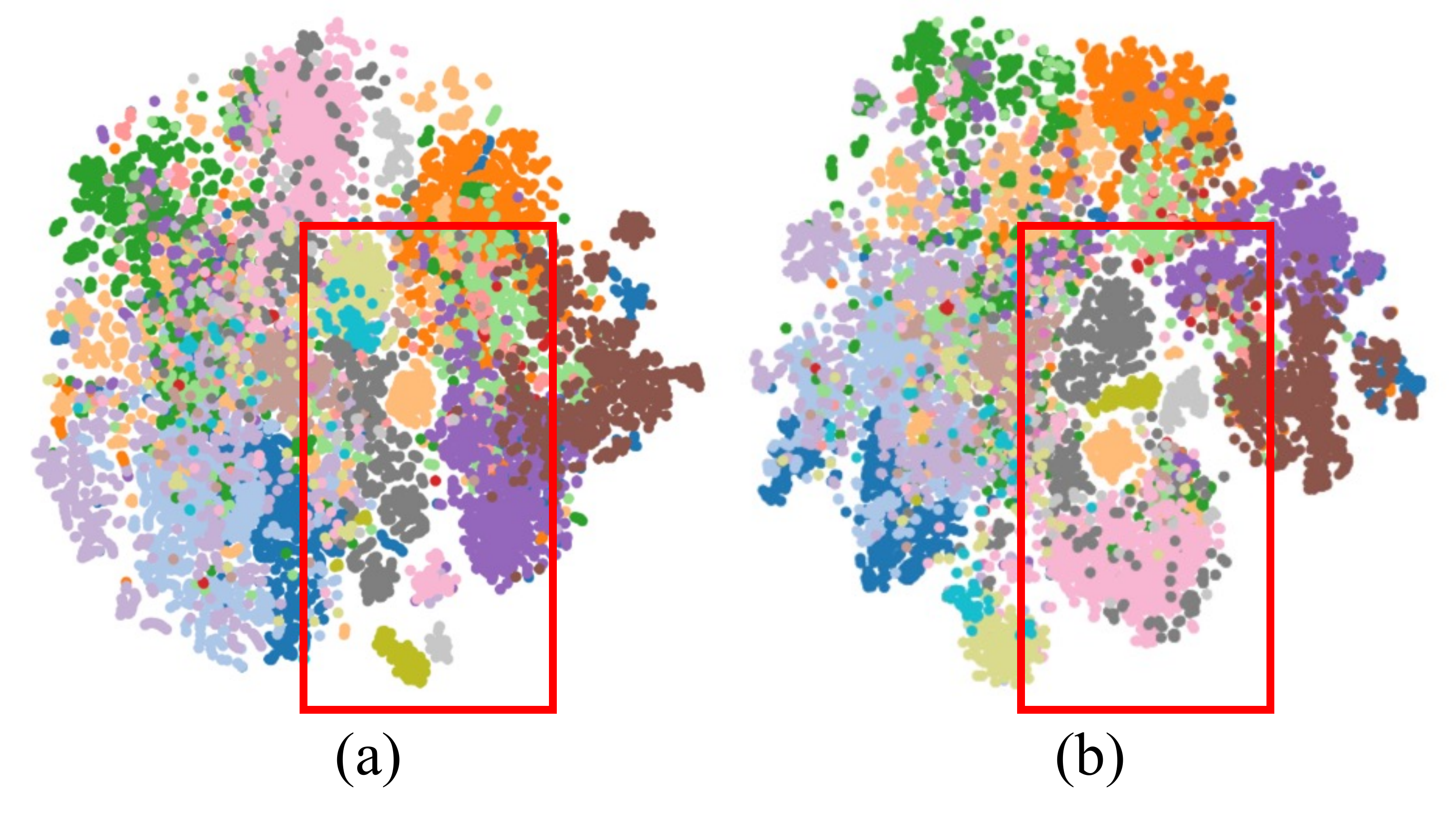}
\vspace{-6pt}
\caption{TSNE visualization of (a) w/o and (b) w/ CFCD.}
\vspace{-16pt}
\label{rebuttal}
\end{figure}

\noindent \textbf{Visualization of class-aware feature distillation.} In Fig.~\ref{rebuttal} below, we provide the TSNE visualization of features with (b) and without (a) our proposed CFCD. Obviously, our class-wise feature distillation makes the models in the cohort achieve better class-wise distinguishing abilities.

\noindent \textbf{Varying backbone models.} 
Since there are two models in our TCC framework, we compare the CPS\cite{chen2021semi} following the two-model structure.
We conduct the ablation studies on the PASCAL VOC 2012 with 1/8 label rate to analyze the impact of the changing backbones, including ResNet, PVT, ConvNext and our TCC.
\textbf{Tab.~\ref{differbackbone} shows that our cohort can better promote the segmentation performance under the dual-student framework in SSL.}  
We fully explore the potential of heterogeneous computing paradigms (MSAs and Convs) via FVCD in the feature space and CCD on the prediction. This way, we successfully introduce ViT into the semi-supervised semantic segmentation. \\
\noindent \textbf{Inference with dual students.} 
During inference, our proposed CFCD makes the two students in the cohort learn from each other, and the ViT-based student achieves better results than the CNN-based one thanks to the better learning ability of MHSAs. Thus, we report the performance of ViT in the cohort in all tables. 
We also report dual students' performance in Tab.~\ref{cnnvitresu}. Obviously, the ViT-based PVT-S in the dual student achieves better results.

\begin{table}[]
    \setlength{\tabcolsep}{4mm}
    \centering
    \caption{TCC with different backbones on PASCAL VOC 2012.}
    \resizebox{\linewidth}{!}{
\begin{tabular}{lccccc}
\toprule
\multirow{2}{*}{Method} & \multirow{2}{*}{Backbones}     & \multicolumn{4}{c}{Label Ratio}                                                                               \\ \cmidrule{3-6} 
                        &                                & 1/16                      & 1/8                       & 1/4                       & 1/2                       \\ \midrule
$U^2$PL                 & \multicolumn{1}{l}{ResNet-101} & \multicolumn{1}{l}{77.21} & \multicolumn{1}{l}{79.01} & \multicolumn{1}{l}{79.30} & \multicolumn{1}{l}{80.50} \\ \midrule
\multirow{5}{*}{Ours}   & ResNet-50                      & 77.79                     & 78.95                     & 80.51                     & 80.98                     \\ 
                        & PVT-M                            & 71.12                     & 73.84                     & 74.44                     & 76.72                     \\ 
                        & ConvNext-S                       & 78.92                     & 80.15                     & 80.81                     & 80.95                     \\ 
                        \cmidrule{2-6} 
                        & TCC-S                          & 79.16                          & 80.28                          & 82.32                            & 82.52                          \\ 
                        & TCC-B                          & \textbf{80.17}                     & \textbf{81.17}                     & \textbf{82.42}                     & \textbf{82.80}                     \\ \bottomrule
\end{tabular}}
\label{differbackbone}
\end{table}
\begin{table}[t!]
    \setlength{\tabcolsep}{3mm}
    \centering
    \caption{Evaluation of CNN and ViT in our TCC-S.}
    \resizebox{\linewidth}{!}{
\begin{tabular}{lccccc}
\toprule
\multirow{2}{*}{Method} & \multirow{2}{*}{Backbone} & \multicolumn{4}{c}{PASCAL VOC} \\ \cmidrule{3-6} 
 &  & 1/16 & 1/8 & 1/4 & 1/2 \\ \midrule
\multirow{2}{*}{TCC-S} & ResNet-50 & 80.98 & 81.34 & 81.78 & 82.67 \\ \cmidrule{2-6} 
 & PVT-S & 81.35 & 83.05 & 83.55 & 84.04 \\ \bottomrule
\end{tabular}}
\label{cnnvitresu}
\end{table}
\section{Conclusion}
In this paper, we proposed TCC, a novel framework for semi-supervised semantic segmentation by exploring the best of both students. Our method subtly incorporates the multi-level consistency regularization on both the predictions and the heterogeneous feature space via pseudo labeling for the unlabeled data. First, the feature variation knowledge is transformed via CF maps between the cohort. Second, we also proposed to distill knowledge from the pixel-wise predictions based on the heterogeneous students. The proposed TCC framework significantly outperformed the SoTA semi-supervised methods by a large margin. 

\noindent \textbf{Limitation and future work:} The proposed TCC demonstrates that using heterogeneous models in consistency regularization gives high-performance gain in semi-supervised learning. For future work, we plan to explore the dual-student framework with more heterogeneous models for semi-supervised semantic segmentation.




\bibliographystyle{IEEEtran}
\bibliography{IEEEexample}

\end{document}